\documentclass[runningheads]{llncs}
\usepackage[T1]{fontenc}
\usepackage{graphicx}
\usepackage{booktabs}

\usepackage{hyperref}
\usepackage{float}
\usepackage{xcolor}
\usepackage{amsmath}
\usepackage{bbm}
\usepackage{tikz}
\usetikzlibrary{shapes.geometric, arrows}
\usetikzlibrary{calc, fit, positioning}
\usepackage{caption}
\usepackage{subcaption}

\usepackage[misc]{ifsym}

\usepackage{mwe}

\begin{document}

\title{A two-step sequential approach for hyperparameter selection in finite context models}

\titlerunning{FCM hyperparameter selection}

\author{José Contente\inst{1}\orcidID{0009-0001-9354-4546} \and
Ana Martins\inst{1,2}\orcidID{0000-0003-4860-7795} \and Armando J. Pinho\inst{1,2}\orcidID{0000-0002-9164-0016} \and Sónia Gouveia\inst{1,2}\orcidID{0000-0002-0375-7610}}

\authorrunning{J. Contente et al.}

\institute{Institute of Electronics and Informatics Engineering of Aveiro (IEETA), \\
Department of Electronics, Telecommunications and Informatics (DETI), \\ 
University of Aveiro (UA), Aveiro, Portugal \email{\{jfcc11, a.r.martins, ap, sonia.gouveia\}@ua.pt}
\and
Intelligent Systems Associate Laboratory (LASI), Portugal}

\maketitle              
\begin{abstract}

Finite-context models (FCMs) are widely used for compressing symbolic sequences such as DNA, where predictive performance depends critically on the context length $k$ and smoothing parameter $\alpha$. In practice, these hyperparameters are typically selected through exhaustive search, which is computationally expensive and scales poorly with model complexity.

This paper proposes a statistically grounded two-step sequential approach for efficient hyperparameter selection in FCMs. The key idea is to decompose the joint optimization problem into two independent stages. First, the context length $k$ is estimated using categorical serial dependence measures, including Cramér’s $\nu$, Cohen’s $\kappa$ and partial mutual information (pami). Second, the smoothing parameter $\alpha$ is estimated via maximum likelihood conditional on the selected context length $k$. Simulation experiments were conducted on synthetic symbolic sequences generated by FCMs across multiple $(k,\alpha)$ configurations, considering a four-letter alphabet and different sample sizes. Results show that the dependence measures are substantially more sensitive to variations in $k$ than in $\alpha$, supporting the sequential estimation strategy. As expected, the accuracy of the hyperparameter estimation improves with increasing sample size. Furthermore, the proposed method achieves compression performance comparable to exhaustive grid search in terms of average bitrate (bits per symbol), while substantially reducing computational cost. Overall, the results on simulated data show that the proposed sequential approach is a practical and computationally efficient alternative to exhaustive hyperparameter tuning in FCMs.

\end{abstract}

\section{Introduction}

Data compression, that is, reducing data digital size by encoding information using fewer bits, is an increasingly important task for efficient storage of information in a fast-moving data-driven world. Furthermore, compression techniques have underlying models that attempt to reproduce as closely as possible the information source to be compressed. Thus, these models are interesting on their own as they can provide insight into the statistical properties of the data. One such model are finite context models (FCMs), widely used in the compression of finite alphabet sequences, such as DNA or proteins \cite{pratas2019,silva2020,silva2021}. Finite context models describe sequences from a finite alphabet, where the probability of observing the next symbol $s$ depends only on the $k$ previous symbols, i.e., the context. This probability is calculated for a fixed $k$ context using the Lidstone estimator \cite{lidstone1920}, where $\alpha$ is a smoothing factor. As such, FCMs are described by hyperparameters context ($k$) and smoothing factor $(\alpha)$. Current approaches to set these hyperparameters in compressors are based on exhaustive trial and error procedures (grid search), which implies that the compression technique must be employed at every trial, so that the ``best'' combination of hyperparameters in the search space is identified. Thus, the wide use of FCMs advocates for a timely manner to find the ``best'' set of hyperparameters, improving compressors time efficiency.

Finite context models are, in fact, discrete-time Markov models, which, from a time series analysis perspective, have a close relation to autoregressive models, i.e., an observation at time $t$ relies on the previous $p$ observations. Thus, concepts commonly used in time series analysis to describe serial dependence are of high relevance in this setting. For real-valued time series, the autocorrelation function (acf) and partial autocorrelation (pacf) play an important role in the study of structural serial dependence \cite{shumway2006}. The acf provides information on the correlation between two time points, whereas pacf provides this information, conditional on intermediate time points, meaning that, it provides a more accurate representation of the serial dependence \cite{shumway2006}. In fact, pacf is used to identify the order $p$ of autoregressive models \cite{shumway2006}. Thus, given the categorical nature of FCMs, to bridge this knowledge into its domain requires the use of counterpart measures adequate for this type of data. The development of serial metrics for categorical time series faces several challenges, but continuous correlation measures such as Cohen's $\kappa$ and Cramer's $\nu$ have been adapted for categorical time series \cite{weiss2008}. Another interesting measure, is the partial auto mutual information (pami, for short) proposed by Biswas and Gua (2009) \cite{biswas2009} to deal with categorical data. The authors have shown that this measure behaves in a similar fashion to the pacf, which supports its use for model order identification. Although, this measure was proposed under a time series perspective, it makes use of the concept of mutual information, thus, relating to information theory and compressors. As such, pami is a very strong contester to aid identifying the  hyperparameter $k$.

The previous metrics allow only the selection of $k$, however, it is also necessary to set $\alpha$ in the Lidstone estimator. This estimator requires that the size of the context ($k$) is previously fixed, which advocates for a sequential strategy where the value of $\alpha$ is chosen after $k$ has been set. Thus, the selection of $\alpha$ can be framed within a principled probabilistic modelling perspective. Specifically, under the Lidstone formulation, the smoothed predictive probability corresponds to the posterior expectation of a multinomial distribution with a symmetric Dirichlet prior. Consequently, estimating $\alpha$ becomes a hyperparameter inference problem. A natural approach is to adopt an empirical Bayes strategy, whereby the multinomial probability vectors associated with each context are treated as latent variables and integrated out, yielding a Dirichlet - multinomial marginal likelihood for the observed count vectors. Pooling information across all contexts of fixed order $k$ enables the estimation, as each context provides an independent multinomial observation contributing to the likelihood of $\alpha$. The resulting estimator is obtained by maximizing the joint marginal likelihood over $\alpha$. This approach  yields a data-driven estimate of $\alpha$, coherent with the probabilistic interpretation of Lidstone smoothing factor, and naturally adapts to the amount of information available in the collection of contexts.

Therefore, the goal of this work is to introduce a two-step sequential approach for the selection of FCM hyperparameters $(k, \alpha)$. The first step of the approach consists in fixing $k$ using categorical serial dependence measures as features. Then, $\alpha$ is estimated via maximum likelihood conditional on the value of $k$. By separating the context length from the smoothing factor, the proposed approach reduces the dimensionality of the optimization problem and avoids the combinatorial search typically required for joint hyperparameter tuning. Hence, this approach will contribute to largely reduce the computational burden and time cost of compression tasks.

The remaining of the paper is outlined as follows: section 2 provides background information finite context models and their hyperparameters, section 3 presents the methods used to develop the two-step sequential approach, including the simulation study design and performance evaluation. Section 4 presents the results of the experimental study and its discussion. Lastly, section 5 is devoted to the main conclusions and future work.

\section{Finite context models and hyperparameters}

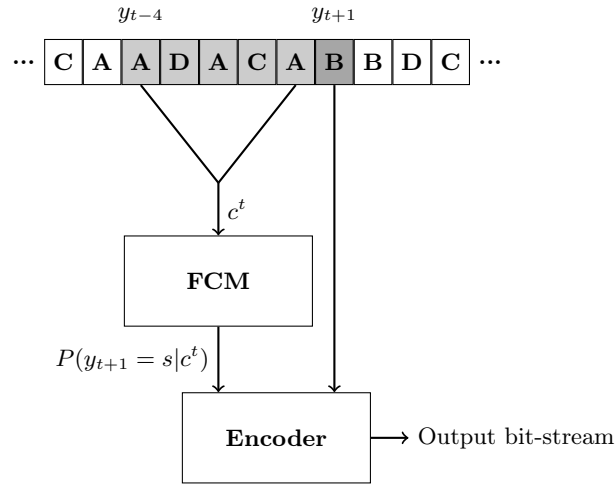
\begin{figure}[ht]
    \centering
    \begin{tikzpicture}[
        box/.style={draw, minimum width=0.5cm, minimum height=0.6cm},
        block/.style={draw, rectangle, minimum width=2.5cm, minimum height=1.2cm},
        arrow/.style={->, thick},
        line/.style = {-, thick}
    ]

    \node[box] (g) {\textbf{C}};
    \node[left=0cm of g.west, anchor=east] (dots1) {\textbf{...}};
    \node[box, right=0cm of g] (a1) {\textbf{A}};
    \node[box, right=0cm of a1, fill=gray!40] (a2) {\textbf{A}};
    \node[box, right=0cm of a2, fill=gray!40] (t) {\textbf{D}};
    \node[box, right=0cm of t, fill=gray!40] (a3) {\textbf{A}};
    \node[box, right=0cm of a3, fill=gray!40] (g2) {\textbf{C}};
    \node[box, right=0cm of g2, fill=gray!40] (a4) {\textbf{A}};
    \node[box, right=0cm of a4, fill=gray!70] (c1) {\textbf{B}};
    \node[box, right=0cm of c1] (c2) {\textbf{B}};
    \node[box, right=0cm of c2] (t2) {\textbf{D}};
    \node[box, right=0cm of t2] (g3) {\textbf{C}};
    \node[right=0cm of g3.east, anchor=west] (dots2) {\textbf{...}};
    
    \node[above=3pt of a2] {$y_{t-4}$};
    \node[above=3pt of c1] {$y_{t+1}$};
    
    \node[block, below=2cm of a3] (fcm) {\textbf{FCM}};
    \node[block] (enc) at ($(a3)!0.5!(c1) + (0,-5cm)$) {\textbf{Encoder}};

    \coordinate (merge) at ([yshift=0.7cm]fcm.north);
    
    \draw[line] (a2.south) -- (merge);
    \draw[line] (a4.south) -- (merge);

    \draw[arrow] (merge) -- node[right] {$c^t$} (fcm.north);

    \draw[arrow] (fcm.south) -- node[left] {$P(y_{t+1}=s|c^t)$} (fcm.south |- enc.north);
    
    \draw[arrow] (c1.south) -- (c1 |- enc.north);
    \draw[arrow] (enc.east) -- ++(0.5,0) node[right] {Output bit-stream};
    \end{tikzpicture}
    \caption{Illustration of the usage of a finite context model in a compression task, showing how the probability of the next outcome, $y_{t+1}$, is conditioned by the last $k$ outcomes ($k= 5$, in this example). Adapted from \cite{pinho2010}. }
    \label{fig:fcm_example}
\end{figure}

Finite context models are Markov models used in the modeling of serial dependence. In this particular case, the interest lies in modeling the serial dependence of a categorical process $Y_t$ described by a finite alphabet $\mathcal{A}$ with range $\{a_1, a_2, \dots, a_r\}$. In this type of model, the occurrence of an observation of a categorical time series $y_t$, where $t=1, \dots, T$, at a time step $t$, depends only on the previous $k$ observations, i.e., the context length. Thus, the hyperparameter $k$ can be referred to as order, context, or even depth of the model \cite{bell1990,sayood2017}. Figure \ref{fig:fcm_example} displays an example of how an FCM works in the compression paradigm. The data sequence is generated from a four-symbol alphabet $\mathcal{A}=\{A,B,C,D\}$. The observation in the time instance $t+1$ relies on the previous $k=5$ observations. Thus, the set of observations $y_{t-4}, ..., y_{t}$, is the conditioning context $c^t$ that allows to calculate the probability of observing a given symbol $s$ at $t+1$. Remark that the number of conditioning states of the model is $|\mathcal{A}|^k$, dictating its complexity. Thus, the hyperparameter $k$ exponentially increases the number of conditioning states for a given alphabet.

The probability that the next outcome equals a given symbol, $y_{t+1} = s$ is obtained using the Lidstone estimator \cite{lidstone1920}
\begin{equation}
    P(y_{t+1} = s|c^t ) = \frac{n_s^t + \alpha}{\sum\limits_{a \in \mathcal{A}} n_a^t + |\mathcal{A}|\alpha},
    \label{eq:estimator_prob}
\end{equation}
where $n_s^t$ represents the number of times that, in the past, the symbol $s$ was generated having $c^t$ as the conditioning context, and $|\mathcal{A}|$ is the cardinality of the alphabet (the cardinality is 4 in Fig. \ref{fig:fcm_example} example). In equation \eqref{eq:estimator_prob}, the factor $\alpha$ controls how much probability is assigned to unseen (but possible) events, playing a key role in the case of high-order models \cite{pinho2011}. The Lidstone estimator reduces to Laplace’s estimator for $\alpha = 1$  \cite{laplace1814}, and to the Jeffreys/ Krichevsky and Trofimov estimator for $\alpha = 1/2$ \cite{jeffreys1946,krichevsky1981}. The smoothing factor $\alpha > 0$, but often values in the range $[0,1]$ are considered where $\alpha = 0$ is the no smoothing case. Thus, full specification of a finite context model is achieved with the hyperparameters $k$ and $\alpha$. 

Additionally, in Fig. \ref{fig:fcm_example} is also represented an arithmetic encoder, which generates an output of bit-streams with average bitrates almost identical to the entropy of the model \cite{bell1990,sayood2017,solomon}. The theoretical average bitrate (entropy) of the finite-context model after encoding $T$ symbols is given by \cite{pinho2010}
\begin{equation}
H_T=-\frac{1}{T} \sum_{t=0}^{T-1} \log _2 P\left(x_{t+1}=s \mid c^t\right) \quad bps,
\label{eq:entropy}
\end{equation}
where $bps$ stands for bits per symbol. Note that the entropy of any four symbol alphabet is, at most, two bps, which is achieved when the symbols are independent and equally likely. Suppose that for the example in Fig. \ref{fig:fcm_example}, with a fixed value for $\alpha$, $P(y_{t+1} = C|c^t ) = 0.1$ and $P(y_{t+1} = A |c^t ) = 0.4$, then the theoretically average bitrate would be $H_1 = 3.32$ and $H_1 = 1.32$, respectively. This shows that for symbol C, given that it is less probable, more than two bits are need for compression. In contrast, for symbol A, less bits would be required. Thus, the arithmetic compressor provides insight on how well the FCM describes the underlying data.




\section{Methods}

Figure \ref{fig: intro_scheme} outlines the two-step sequential framework employed to obtain the pair of optimal hyperparameters ($k^* ,\alpha^*$). First, from an observed categorical time series, $y_t$, feature extraction is performed based on pami and other serial dependence metrics. Then $k^*$ is chosen by identificatying the lag at which the maximum serial dependence occurs. This step identifies the order of the FCM that best captures the serial dependence structure of the sequence, independently of any smoothing assumptions. The second-step of the strategy is fed with both $y_t$ and $k^*$, and the smoothing parameter $\alpha^*$ is estimated via maximum likelihood conditional on $k^*$. At the end of the procedure ,a pair of optimal hyperparameters is obtained. Thus, rather than relying on exhaustive joint search procedures, the proposed approach decomposes the problem into two stages guaranteeing its computational efficiency.

\begin{figure}[H]
    \centering
    \begin{tikzpicture}[
        box/.style = {rectangle, draw, rounded corners, minimum width=2cm, minimum height=1cm},
        arrow/.style = {->, thick},
        line/.style = {-, thick}
    ]

    \node (box2) [box, align=center] {$k^*$ Selection};
    \node (box4) [box, left of=box2, node distance=2.9cm] {Feature Extration};

    \node (big1) [draw, thick, rounded corners=8pt,fit=(box2)(box4), inner sep=0.4cm] {};
    \node[above=3pt of big1]{\textbf{Compute $k^*$}};

    \node (box1) [left=0.4cm of big1.west, anchor=east] {$y_t$};

    \node (box8) [ right of=big1, align=center, node distance=3.6cm] {$k^*$};
    \node (box5) [box, right of=box8, node distance=1.8cm] {ML Estimation};

    \node[above=3pt of box5] {\textbf{Compute $\alpha^*|k^*$}};

    \node (box6) [ right of=box5, node distance=2.2cm] {$(k^*, \alpha^*)$};

    \draw [arrow] (box1) -- (big1);
    \draw [arrow] (box4) -- (box2);
    \draw [line] (big1) -- (box8);

    \draw[arrow] (box1.south) |- ([yshift=-0.4cm]big1.south) -| ([yshift=-7pt]box8.south) |-([yshift=-7pt]box5.west);

    \draw [arrow] (box8) -- (box5);
    \draw [arrow] (box5) -- (box6);

    \end{tikzpicture}
    \caption{Outline of the two-step sequential strategy for hyperparameter context ($k^*$) and smoothing factor ($\alpha^*$) selection.}
    \label{fig: intro_scheme}
\end{figure}

In the following subsections, the selection of $k^*$ via pami, and the computation of $\alpha^*$ conditionally on $k^*$ are described. Moreover, other categorical time-series features employed for context length selection are also discussed. Then, the simulation study design and performance metrics are presented.




\subsection{Determination of $k^*$ via pami}

The partial auto mutual information (pami) is a special case of the conditional mutual information ($\operatorname{I}$). The latter measures how much information two random variables share, when accounting for the effect of a third one, i.e.,
\begin{equation}
    \operatorname{I}(X_1,X_2|Z) = E\left( \operatorname{log}\frac{\operatorname{P}(X_1,X_2|Z)}{\operatorname{P}(X_1|Z)P(X_2|Z)}\right),
    \label{eq:cond_mi}
\end{equation}
where $\operatorname{P}(.|Z)$ are conditional probabilities. 
The derivation of a time-lagged version of \eqref{eq:cond_mi}, results in pami, where $X_1$ and $X_2$ are replaced by $Y_t$ and $Y_{t+h}$, and $Z$ are now the time lags between in between these these time-points, i.e., $\mathcal{F}_t = Y_{t+1}, \dots, Y_{t+h-1}$ \cite{biswas2009}. Therefore,
\begin{equation}
    \operatorname{pami}(h) = \operatorname{E}\left(\operatorname{log}\frac{\operatorname{P}(Y_t, Y_{t+h}| \mathcal{F}_t)}{\operatorname{P}(Y_t| \mathcal{F}_t) \operatorname{P}(Y_{t+h}| \mathcal{F}_t)} \right)
\end{equation}
The similarity of pami expression to the pacf one is striking, in fact, \cite{biswas2009} showed that for a discrete AR($p$) process, $\operatorname{pami}(h) = 0$ for $h>p$ supporting its use for model order selection.

After computing the pami for a data sequence, a criterion is needed to select the optimal value $k^*$. A maximum-based criteria is defined, i.e., the lag at which the maximum pami value is observed is set as the optimal $k^*$.

\subsection{Determination of $\alpha^* \mid k^*$}

The Lidstone estimator \eqref{eq:estimator_prob} corresponds to the posterior expectation of the multinomial probabilities under a symmetric Dirichlet prior $\text{Dirichlet}(\alpha,\ldots,\alpha)$ \cite{bishop2006,gelman2013}. Consequently, the estimation of $\alpha$ can be formulated as a hyperparameter inference problem. 

Let $C_{k^*}$ denote the set of contexts of order $k^*$ extracted from the sequence. For each context $c \in C_{k^*}$, let
\begin{equation}
    \mathbf{n}_c = (n_{c,1}, \ldots, n_{c,|\mathcal{A}|})
\end{equation}
denote the multinomial count vector associated with that context, with total count
$N_c = \sum_{s \in \mathcal{A}} n_{c,s}$. Under the symmetric Dirichlet prior, integrating out the multinomial parameters yields the Dirichlet--multinomial marginal likelihood

\begin{equation}
p(\mathbf{n}_c \mid \alpha) =
\frac{\Gamma(|\mathcal{A}|\alpha)}
{\Gamma(N_c + |\mathcal{A}|\alpha)}
\prod_{s \in \mathcal{A}}
\frac{\Gamma(n_{c,s} + \alpha)}
{\Gamma(\alpha)}.
\label{eq:dm}
\end{equation}

Assuming conditional independence across contexts, the joint log-marginal likelihood over all contexts of order $k^*$ is given by
\begin{equation}
\ell(\alpha)
  = \log \prod_{g=1}^G p\bigl(\mathbf{n}^{(g)} \mid \alpha\bigr),
\end{equation}
and the empirical Bayes estimate of $\alpha$ is defined by
\begin{equation}
\alpha^* =
\arg\max_{\alpha > 0} \ell(\alpha).
\label{eq:alpha_est}
\end{equation}

Because this optimisation problem is one-dimensional, the maximum can be obtained efficiently using numerical optimisation methods such as Newton--Raphson or gradient-based search. 

\subsection{Other candidate features}

The additional features explored for the hyperparameter $k^*$ 
are conventional categorical time series dependence measures, namely Cramer's $\upsilon$ and Cohen's $\kappa$. These serial dependence metrics were adapted from continuous data, to handle the discrete nature of categorical time series \cite{weiss2008}. 
Let $p_i = \operatorname{P}(Y_t = i)$ be the marginal probability of the $i$th category of $Y_t$, obtained with the relative frequencies estimator 
\begin{equation}
\hat{p}_i = \frac{1}{T} \sum_{t=1}^{T} \mathbbm{1}_{i}(Y_t) ,
\end{equation}
where $\mathbbm{1}_{i}()$ is the indicator function with $\mathbbm{1}_i (Y_t) = 1$ if $Y_t = i$ and $0$ otherwise. Moreover, consider the notation for the joint probability $p_{ij}(h) = \operatorname{P}(Y_t = i, Y_{t-h}=j)$, with $i,j \in \mathcal{A}$, estimated by
\begin{equation}
\hat{p}_{ij}(h) = \frac{1}{T-k}\sum_{t=h+1}^T \mathbbm{1}_{i}(Y_{t}) \mathbbm{1}_{j}(Y_{t-h}).
\end{equation}
Then, Cramer's $\upsilon$, a measure of unsigned serial dependence, i.e. unorientated association, can be defined as \cite{weiss2008}
\begin{equation}
    \upsilon(h) = \sqrt{\frac{1}{r-1}}\sum_{i,j=1}^r \frac{(p_{ij}(h)-p_ip_j)^2}{p_ip_j},
\end{equation}
In contrast, Cohen's $\kappa$ is a signed (orientated) association measure \cite{weiss2008}
\begin{equation}
    \kappa(h) = \frac{\sum\limits_{i=1}^r\big(p_{ii}(h) - p_i^2\big)}{1- \sum\limits_{i=1}^r p_i^2}.
\end{equation}





\subsection{Simulation Study \& Performance Evaluation}

To evaluate the proposed sequential hyperparameter selection strategy, a simulation study was conducted. Synthetic categorical time series were generated from finite-context models (FCMs) defined over a four-symbol alphabet $\mathcal{A} = \{A, B, C, D\}$, $k \in \{1, \cdots, 10\}$ and $\alpha$ defined on a grid of 201 equally spaced values in the interval $[0,1]$. Thus, a total of 2010 combinations of $(k, \alpha)$ were used as the true data-generating process. For each combination, 100 sequences of length $T=100,000$ were generated. Then, pami and the other serial dependence measures were computed for each one of the 201,000 data sequences, to assess their ability in identifying $k^*$.

In a second experiment, random pairs of $(k, \alpha)$ were randomly selected from the 2010 possible combinations to generate 1000 sequences of length $T=\{1,000;$ 
$10,000; 100,000\}$, resulting in 622 unique pairs of ($k, \alpha$). For each sequence, the proposed two-step approach was applied, i.e., $k^*$ selection based on pami, followed by estimation of $\alpha^*$ via maximum likelihood conditional on $k^*$. The sequences were generated from different lengths to evaluate whether the procedure degrades when limits information is available.

For the second experiments with 1,000 instances, performance was evaluated from two complementary perspectives. First, the predictive ability of pami was assessed by comparing the proportion of correctly estimated $k^*$ with the data-generating $k$. In addition, the estimation of $\alpha^*$ was evaluated conditional on $k$ and $k^*$, to assess how the estimation of $\alpha^*$ is impacted by $k^*$. To assess how $\alpha^*|k$ and $\alpha^*|k^*$ are related to $\alpha$, the Pearson correlation was computed,
\begin{equation}
    r(z_i, \alpha) = \frac{\sum_{i=1}^n\left(z_i-\bar{z}\right)\left(\alpha_i-\bar{\alpha}\right)}{\sqrt{\sum_{i=1}^n\left(z_i-\bar{z}\right)^2} \sqrt{\sum_{i=1}^n\left(\alpha_i-\bar{\alpha}\right)^2}},
\end{equation}
where $\alpha_i$ is the data generating parameter, $\bar{\alpha}$ is its average and, $z_i$ can be replaced by $\alpha_i^*|k_i$ or $\alpha_i^*|k_i^*$, for $i=1\dots, \;1,000$ sequences. 

Second, the practical impact of the estimated hyperparameters on compression performance was evaluated using the theoretical average bitrate (bits per symbol) defined in equation (\ref{eq:entropy}). For each sequence, the bitrate obtained using the pair of estimated hyperparameters $(k^*, \alpha^*)$ was compared with the bitrate obtained via exhaustive grid search. This procedure was carried out for $k \in \{1, \ldots,10\}$ and $\alpha \in \{0,0.1,\ldots,1\}$, yielding a total of 1,010 combinations. Then, the pair $(k,\alpha)$, achieving the minimum bitrate was selected as the optimal grid search configuration. Thus, a fairer comparison is rendered since, in practice, the hyperparameters $(k, \alpha)$ are not known and this is the usual procedure to find them. 


\section{Results}

\begin{figure}[H]
    \centering
    \begin{subfigure}[b]{0.49\textwidth}
        \centering
        \caption*{$(k, \alpha) = (3,0)$}
        \includegraphics[width=\linewidth]{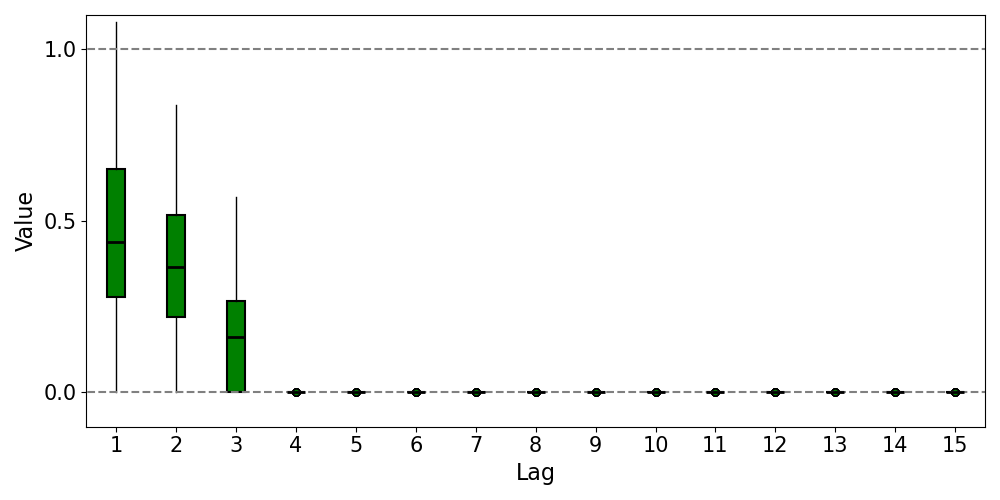}
    \end{subfigure}
    \begin{subfigure}[b]{0.49\textwidth}
        \centering
        \caption*{$(k, \alpha) = (8,0)$}
        \includegraphics[width=\linewidth]{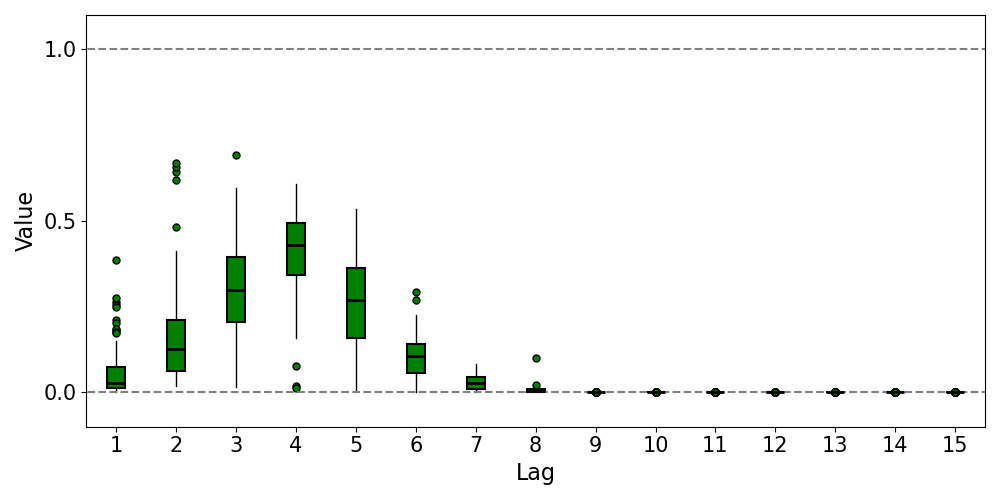}
    \end{subfigure}
    \begin{subfigure}[b]{0.49\textwidth}
        \centering
        \caption*{$(k, \alpha) = (3,0.1)$}
        \includegraphics[width=\linewidth]{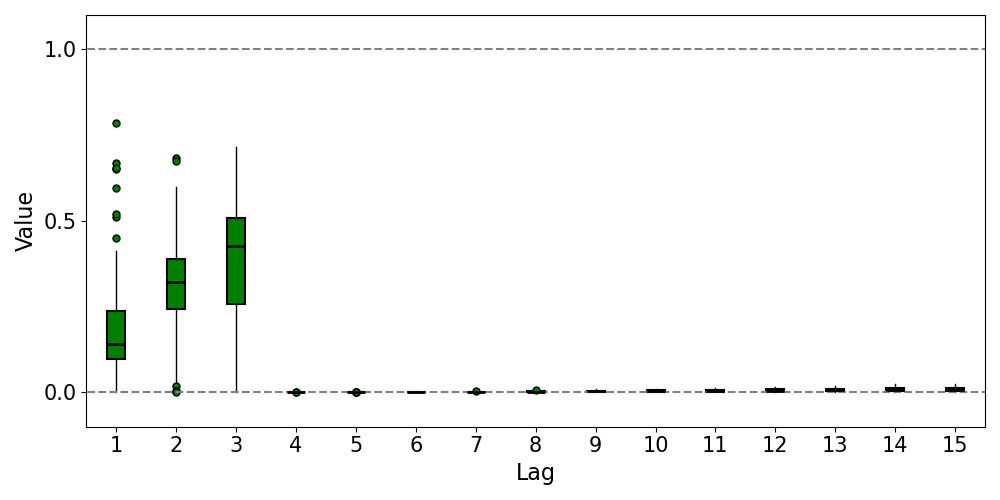}
    \end{subfigure}
    \begin{subfigure}[b]{0.49\textwidth}
        \centering
        \caption*{$(k, \alpha) = (8,0.1)$}
        \includegraphics[width=\linewidth]{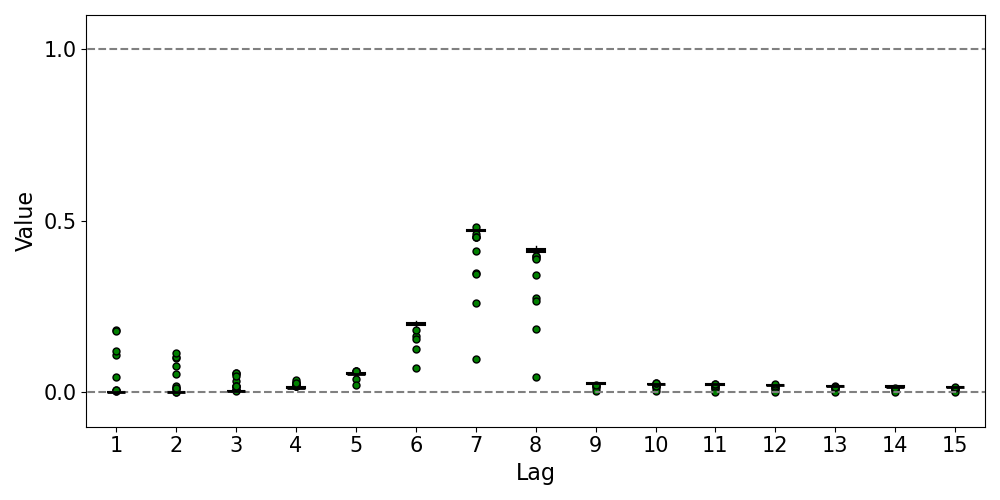}
    \end{subfigure}
    \begin{subfigure}[b]{0.49\textwidth}
        \centering
        \caption*{$(k, \alpha) = (3,0.5)$}
        \includegraphics[width=\linewidth]{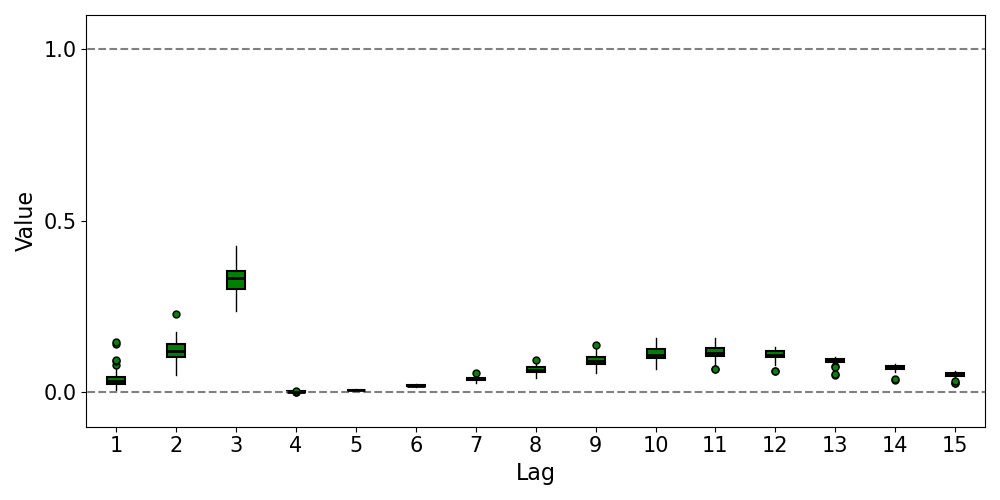}
    \end{subfigure}
    \begin{subfigure}[b]{0.49\textwidth}
        \centering
        \caption*{$(k, \alpha) = (8,0.5)$}
        \includegraphics[width=\linewidth]{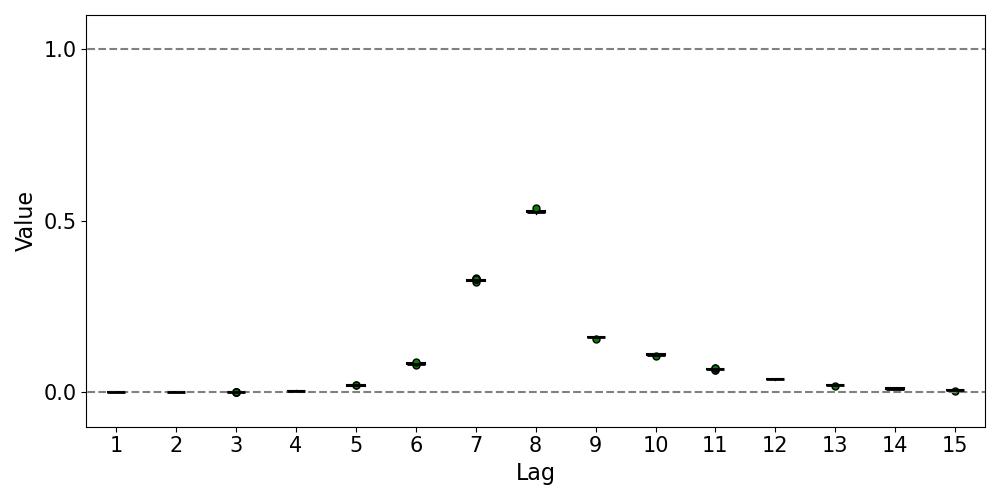}
    \end{subfigure}
    \begin{subfigure}[b]{0.49\textwidth}
        \centering
        \caption*{$(k, \alpha) = (3,0.8)$}
        \includegraphics[width=\linewidth]{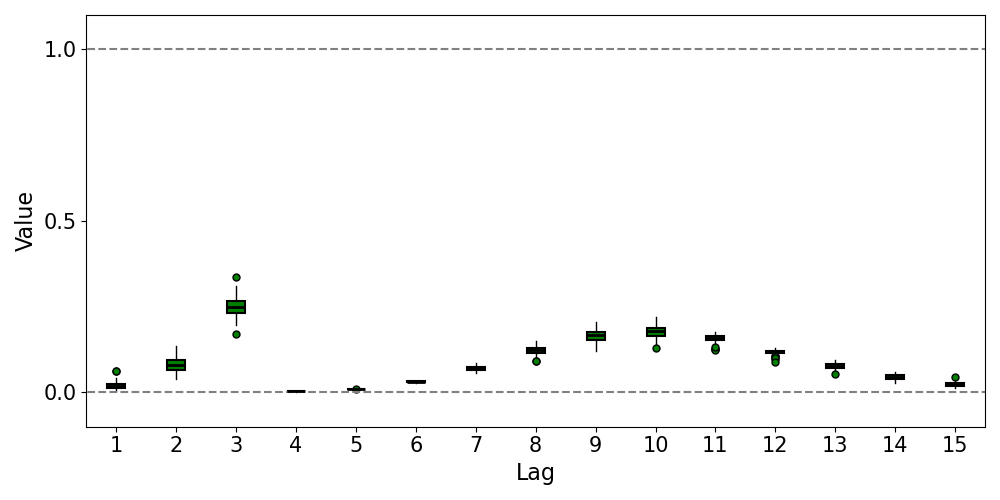}
    \end{subfigure}
    \begin{subfigure}[b]{0.49\textwidth}
        \centering
        \caption*{$(k, \alpha) = (8,0.8)$}
        \includegraphics[width=\linewidth]{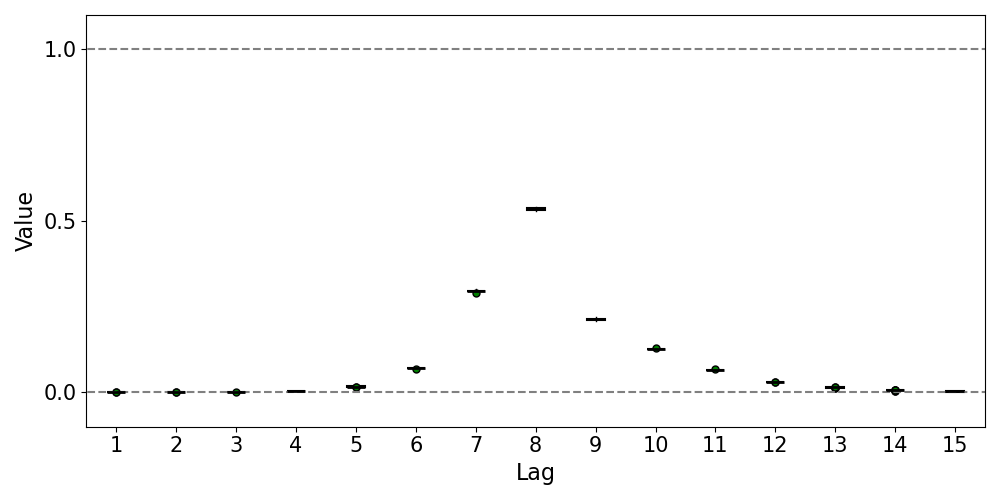}
    \end{subfigure}
    \begin{subfigure}[b]{0.49\textwidth}
        \centering
        \caption*{$(k, \alpha) = (3,1)$}
        \includegraphics[width=\linewidth]{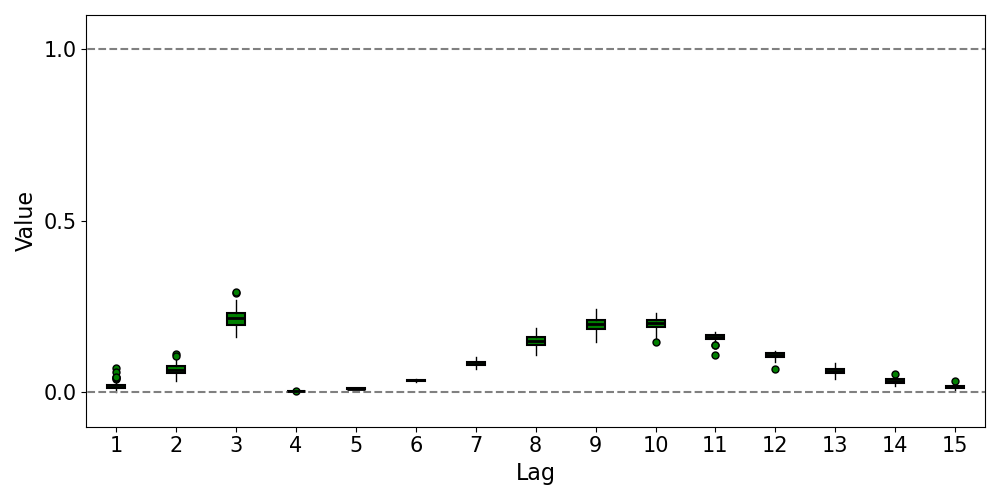}
    \end{subfigure}
    \begin{subfigure}[b]{0.49\textwidth}
        \centering
        \caption*{$(k, \alpha) = (8,1)$}
        \includegraphics[width=\linewidth]{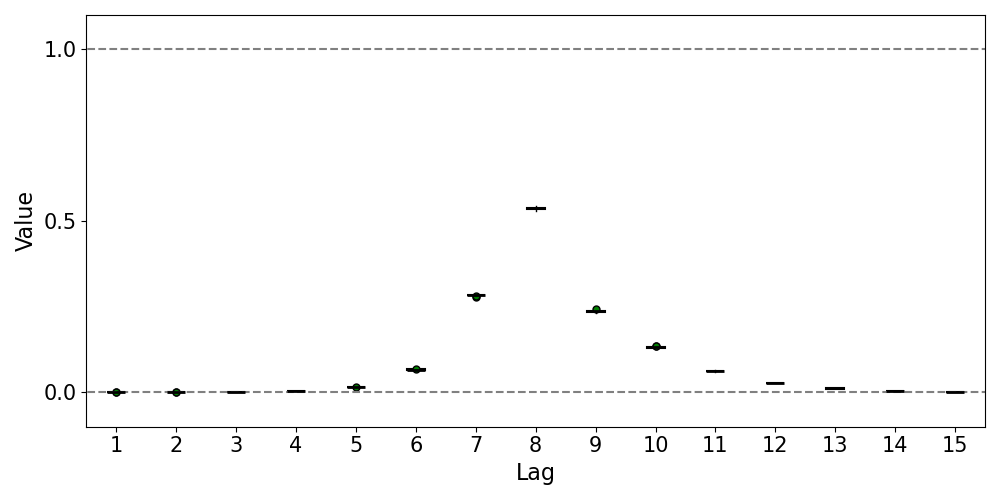}
    \end{subfigure}
    \caption{Boxplots of the distribution of pami for synthetic time series of length 100,000, generated with $k \in \{3,  8\}$  and $\alpha \in \{0, 0.1, 0.5,0.8, 1 \}$.}
    \label{fig:PAMI_100000}
\end{figure}

Figure \ref{fig:PAMI_100000} shows the distribution of pami for the synthetic sequences of length $T=100,000$ for different combinations of $(k, \alpha)$. Two representative context lengths ($k=3$ and $k=8$) are shown for several values of $\alpha$. The results show a clear and consistent pattern, with pami exhibiting a pronounced peak at the lag corresponding to $k$, followed by a rapid decay for higher lags. This behavior is similar to that of the pacf of autoregressive models. Although some variation can be observed as a result of $\alpha$, the pami pattern presents similar shape for $k$. The shape found for $k=3$ for $\alpha > 0.5$ is noteworthy. Larger values of $\alpha$ mean that more weight is given in the Lidstone estimator \eqref{eq:estimator_prob} to the uniform distribution. Thus, this curve maybe explained by giving more weight to less frequent events. This effect is somewhat observed for $k=8$, but overlapped by an effect associated with the serial dependence. Also, remark that pami considers the relative frequency estimator of probabilities, i.e. $\alpha=0$.
Overall, the results suggests that pami is driven mainly by $k$, rather than the smoothing factor ($\alpha$), thus supporting its use in identifying $k^*$. 




\begin{figure}[H]
    \centering
    \begin{subfigure}[b]{0.49\textwidth}
        \centering
        \caption*{$(k, \alpha) = (3,0.1)$}
        \includegraphics[width=\linewidth]{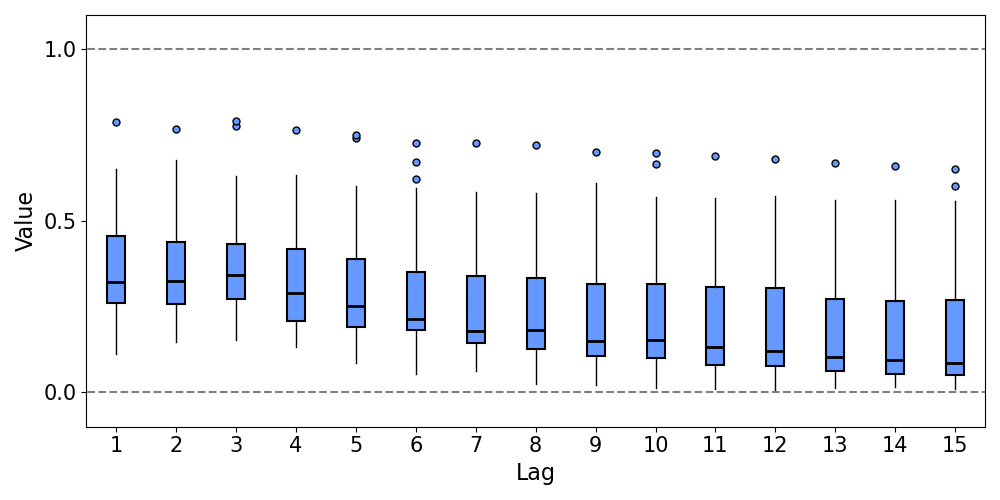}
    \end{subfigure}
    \begin{subfigure}[b]{0.49\textwidth}
        \centering
        \caption*{$(k, \alpha) = (3,0.1)$}
        \includegraphics[width=\linewidth]{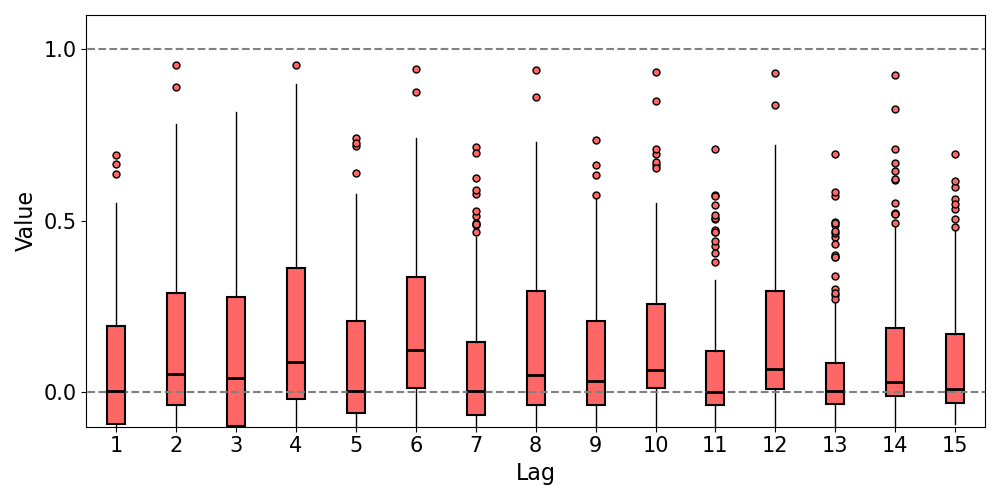}
    \end{subfigure}
    \begin{subfigure}[b]{0.49\textwidth}
        \centering
        \caption*{$(k, \alpha) = (3,0.5)$}
        \includegraphics[width=\linewidth]{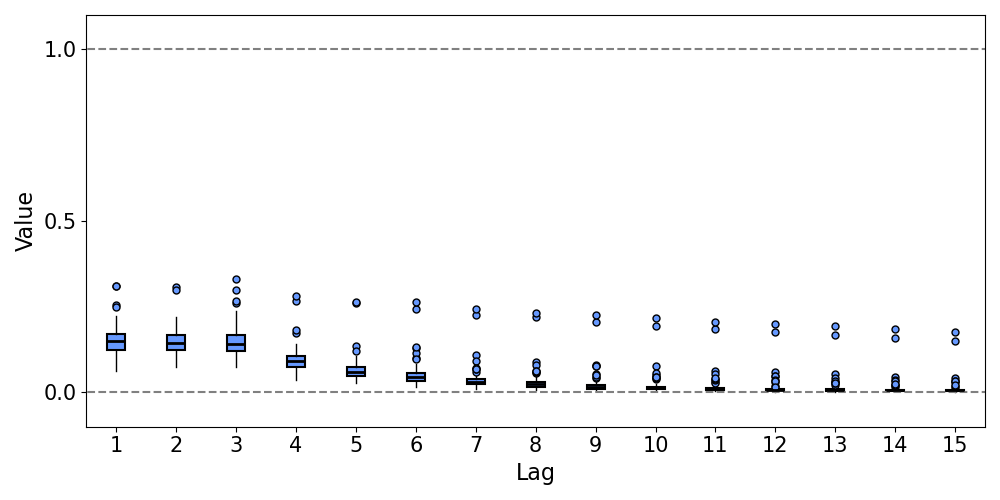}
    \end{subfigure}
    \begin{subfigure}[b]{0.49\textwidth}
        \centering
        \caption*{$(k, \alpha) = (3,0.5)$}
        \includegraphics[width=\linewidth]{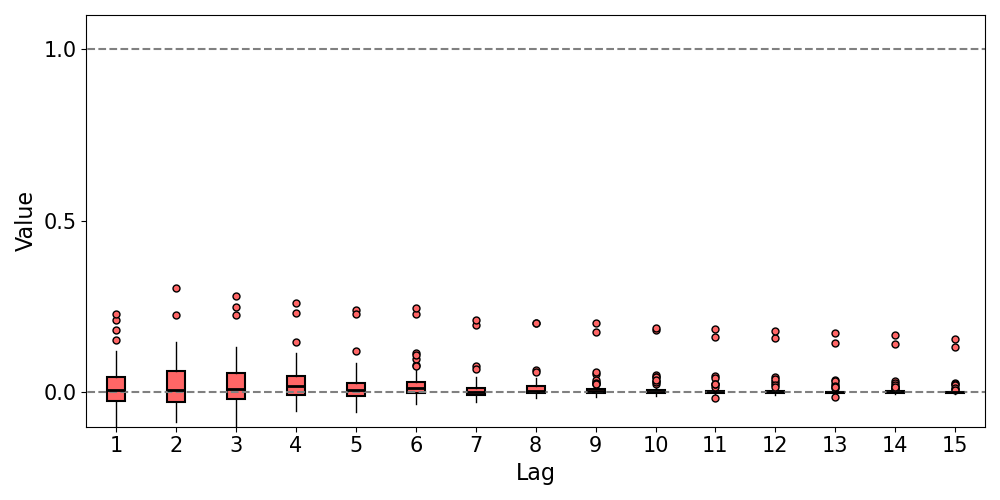}
    \end{subfigure}
    \begin{subfigure}[b]{0.49\textwidth}
        \centering
        \caption*{$(k, \alpha) = (3,0.8)$}
        \includegraphics[width=\linewidth]{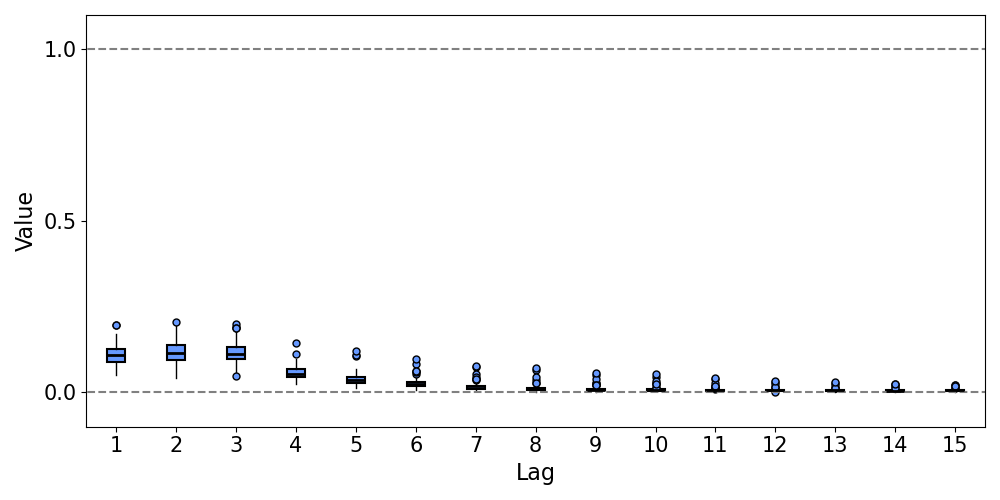}
    \end{subfigure}
    \begin{subfigure}[b]{0.49\textwidth}
        \centering
        \caption*{$(k, \alpha) = (3,0.8)$}
        \includegraphics[width=\linewidth]{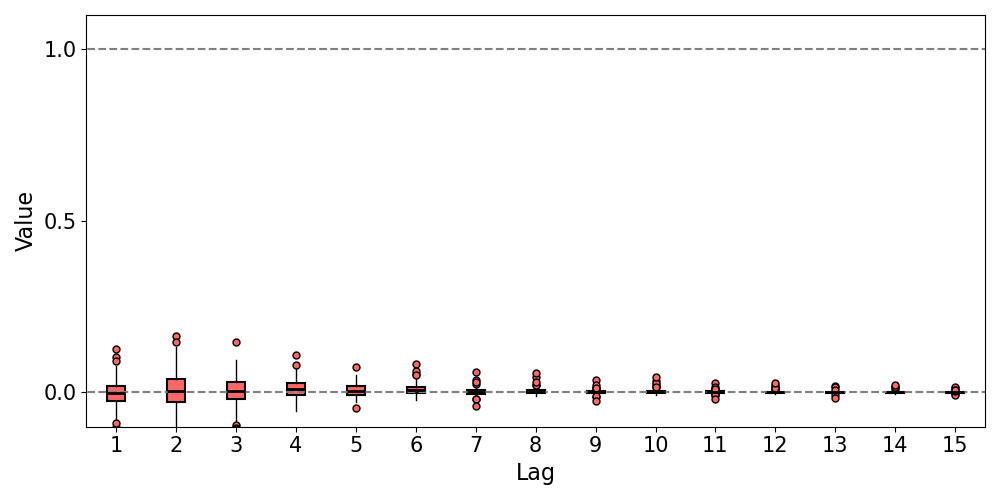}
    \end{subfigure}
    \caption{Boxplots of the distribution of Cramér’s $\nu$ (blue) and Cohen’s $\kappa$ (red) for the synthetic time series of length 100,000, with $k = 3$ and $\alpha \in \{ 0.1,0.5,0.8\}$.}
    \label{fig:cr_co_100000}
\end{figure}

Figure \ref{fig:cr_co_100000} shows the behavior of Cramér’s $\nu$ (blue) and Cohen’s $\kappa$ (red), for sequences generated with $k=3$ and several smoothing factors. For Cramér's, $\nu$ slightly higher values for $k$ up to lag 3, decreasing from there on, can be detected. However, for larger values of $k$, the observed Cramér's $\nu$ pattern rapidly vanishes, with values close to zero for all lags (data not shown). Thus, Cramér’s $\nu$ is not a good option as feature to describe $k$, since for larger $k$ it loses its (small) discriminative ability. Regarding,  Cohen's $\kappa$ no pattern is identified. Thus, these metrics are not helpful for aiding in $k$ selection. 


\begin{figure}[H]
    \centering
    \begin{subfigure}[b]{0.49\textwidth}
        \centering
        \caption*{$(k,\alpha) = (3,0.96)$}
        \includegraphics[width=\linewidth]{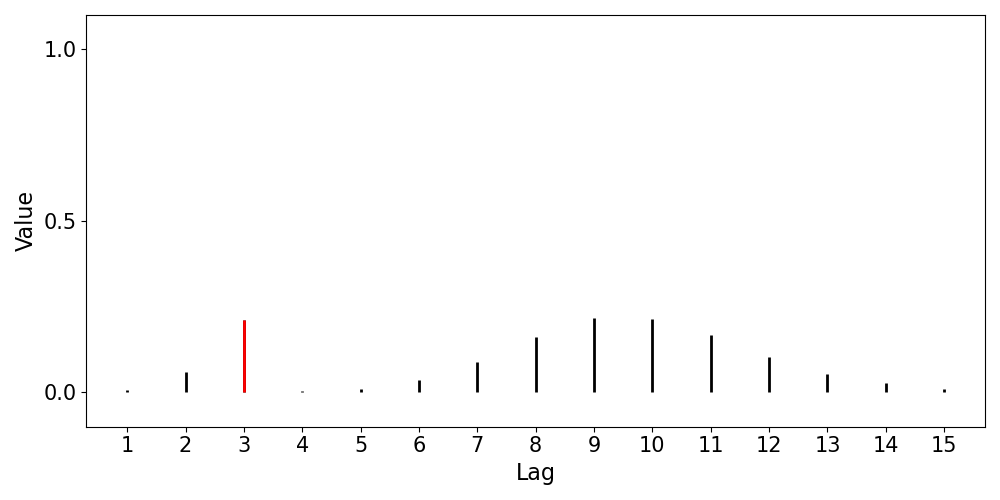}
    \end{subfigure}
    \begin{subfigure}[b]{0.49\textwidth}
        \centering
        \caption*{$(k, \alpha) = (8, 0.22)$}
        \includegraphics[width=\linewidth]{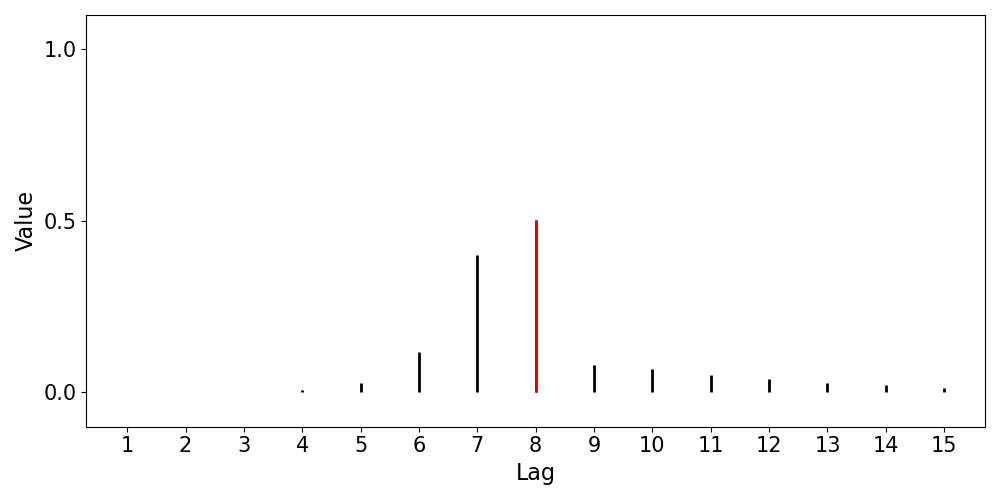}
    \end{subfigure}
    \caption{Partial auto mutual information for two data sequences with $T=100,000$. Maximum value highlighted in red.}
    \label{fig:PAMI_pred}
\end{figure}

Figure \ref{fig:PAMI_pred} highlights the use of the maximum-based pami criteria computed for two data sequences with different values of $(k, \alpha)$. The optimal $k^*$ is chosen as the lag at which the maximum pami value is observed. In both cases, $k^*$ corresponds to the true $k$, although the peak is more evident for $k=8$ (highlighted in red), since for $k=3$ a similar pami value can be found at lag 10.




\begin{figure}[H]
    \centering
    \includegraphics[width=0.99\linewidth]{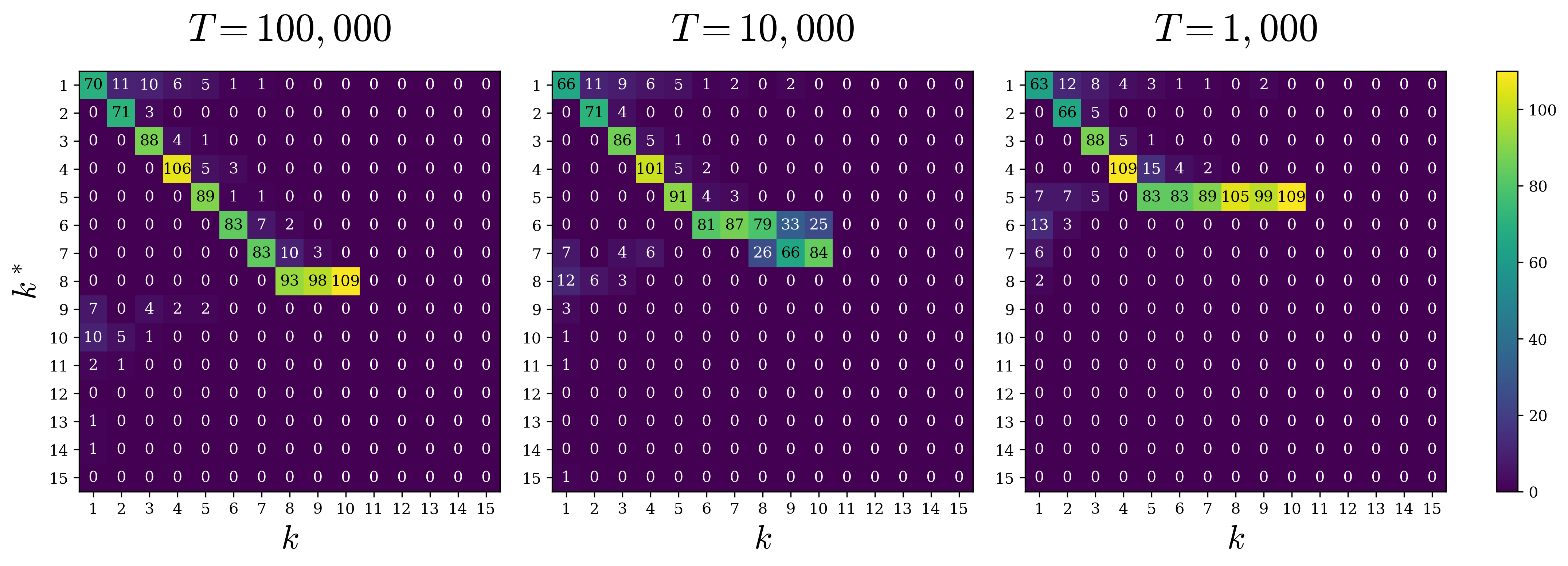}
    \caption{Confusion matrix comparing the context $k$ used to generate the synthetic sequences and the optimal context $k^*$ obtained using the pami-based selection rule for different values of $T$.}
    \label{fig:conf_mat_k}
\end{figure}

After evaluating pami for an exhaustive number of $(k, \alpha)$ values, the assessment of the proposed procedure is performed for the second experiment setting with 1,000 data sequences for the 622 unique pairs of $(k, \alpha)$ and for varying $T$ lengths. First, confusion matrices were built using the maximum-based pami criterion to select $k^*$. The matrix compares the true context ($k$) with the values obtained by the proposed method ($k^*$) for $T=\{=1,000; 10,000; 100,000\}$ (Fig. \ref{fig:conf_mat_k}). The diagonal identifies the number of correctly predicted $k^*$ values. For $T=100,000$ nearly 70\% of the series had $k^*$ correctly predicted. This value is limited as a result of the large miss-classification for $k=9,10$, which are mainly predicted as $k^* = 8$. For smaller sample sizes, the correctly predicted $k^*$ values decrease to about 50\% for $T=10,000$ and just $40\%$ for $T=1,000$. Moreover, the miss-classification now occurs mostly for values of $k^*=6,7$ and $k^*=5$ for $T= 10,000$ and $T=1,000$, respectively. Thus, this suggests, that the ability of this criteria in identifying $k$ decreases for smaller $T$ and, may be limited by a relation between the context length $k$ and the sample size $T$.

\begin{figure}[h]
    \centering
    \begin{subfigure}[b]{0.325\linewidth}
        \centering
        \caption*{$(\alpha^* | k^*) \quad T= 100,000$}
        \includegraphics[width=\linewidth]{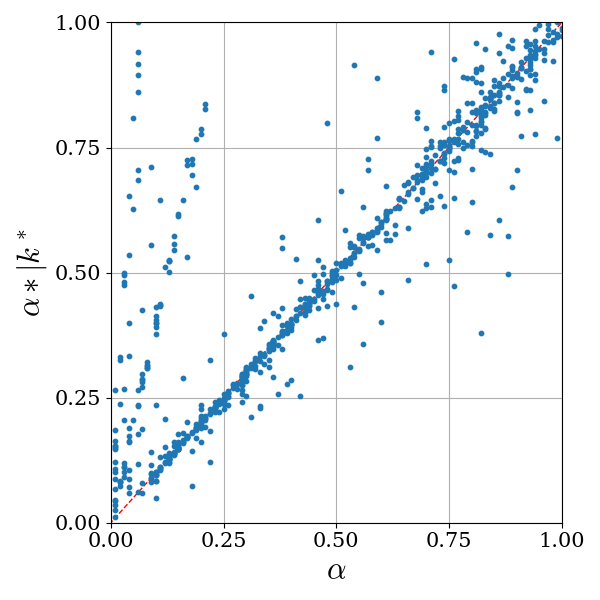}
    \end{subfigure}
    \begin{subfigure}[b]{0.325\linewidth}
        \centering
        \caption*{$(\alpha^* | k^*) \quad T= 10,000$}
        \includegraphics[width=\linewidth]{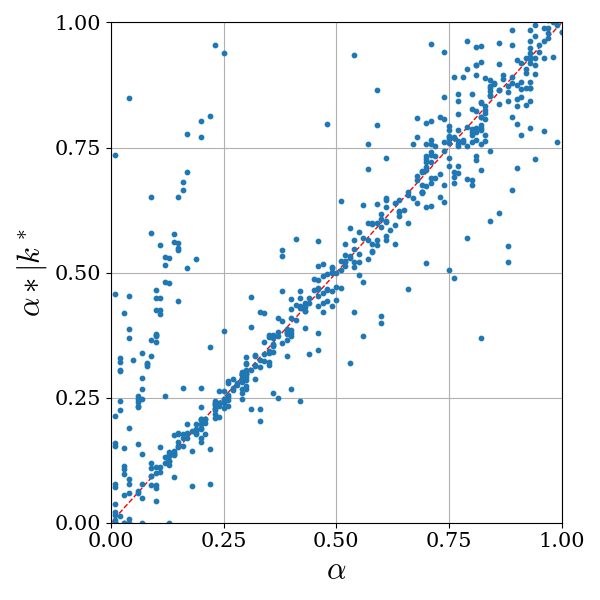}
    \end{subfigure}
    \begin{subfigure}[b]{0.325\linewidth}
        \centering
        \caption*{$(\alpha^* | k^*) \quad T= 1,000$}
        \includegraphics[width=\linewidth]{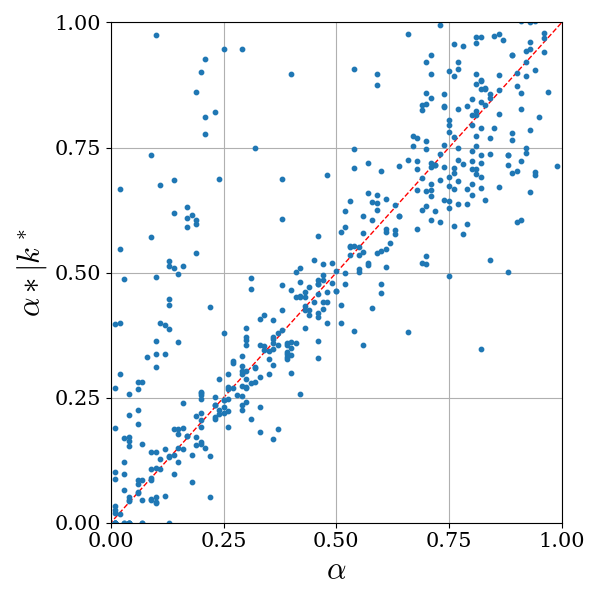}
    \end{subfigure}
    \begin{subfigure}[b]{0.325\linewidth}
        \centering
        \caption*{$(\alpha^* | k) \quad T= 100,000$}
        \includegraphics[width=\linewidth]{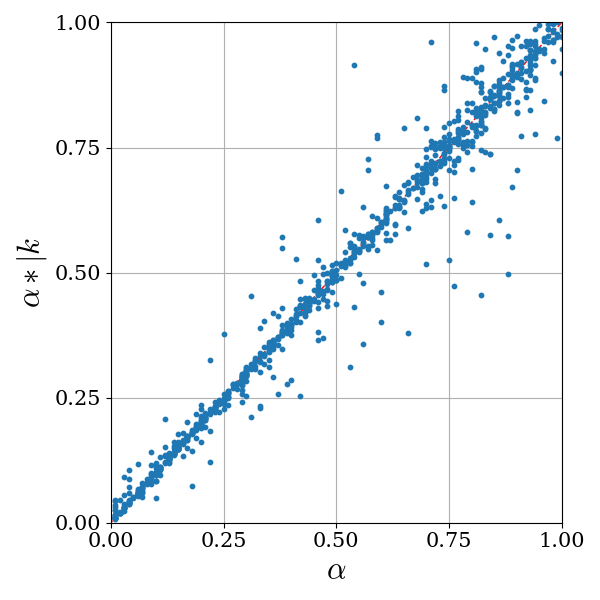}
    \end{subfigure}
    \begin{subfigure}[b]{0.325\linewidth}
        \centering
        \caption*{$(\alpha^* | k) \quad T= 10,000$}
        \includegraphics[width=\linewidth]{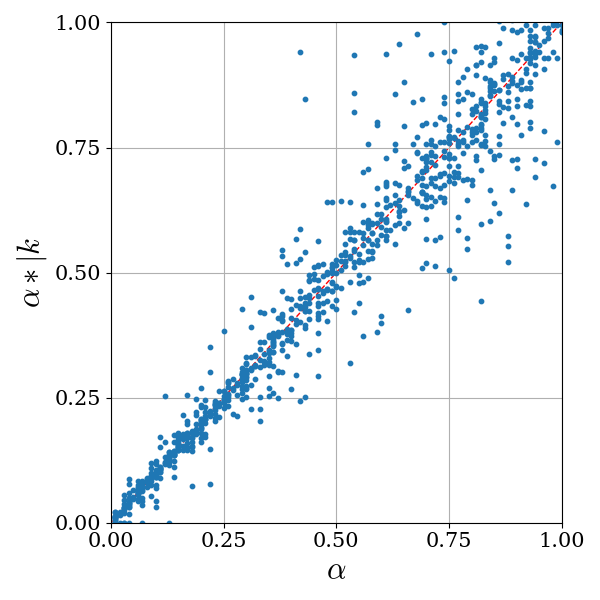}
    \end{subfigure}
    \begin{subfigure}[b]{0.325\linewidth}
        \centering
        \caption*{$(\alpha^* | k) \quad T= 1,000$}
        \includegraphics[width=\linewidth]{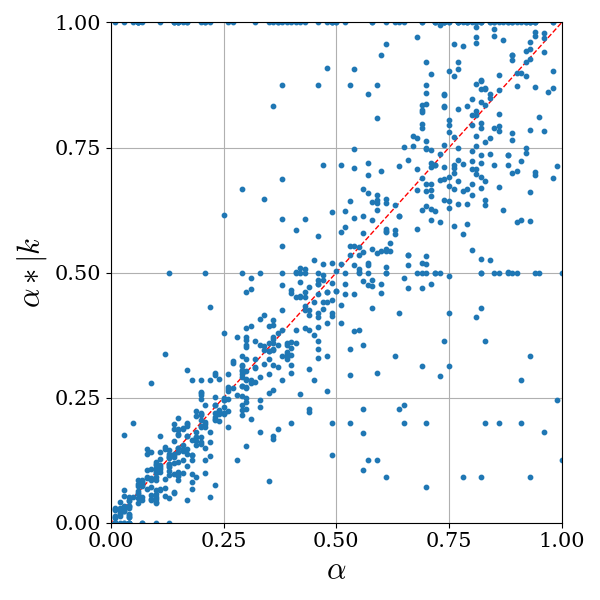}
    \end{subfigure}
    \caption{Dispersion plots of the estimated smoothing factor conditioned to the optimal context $(\alpha^*|k^*)$, and conditioned  to the data-generating context $(\alpha^*|k)$, against the smoothing factor $(\alpha)$ for different sample sizes $T$. Red line is $r=1$.}
    \label{fig:dispersion}
\end{figure}

Figure \ref{fig:dispersion} displays the dispersion plots of 
$(\alpha^*|k^*)$ and $(\alpha^*|k)$, against $\alpha$ used in the data-generating process for different sample sizes $T$, where the red line indicates perfect correlation ($r=1$). The optimization procedure allows $\alpha > 1$, but, for readability, plots are shown for the interval $[0,1]$. When $\alpha^*$ is estimated conditionally on $k^*$, although a considerable amount of points is over the line of perfect correlation, there is still some variability for $T=100,000$. 
As $T$ decreases, varibility increases and the number of points over the $r=1$ line visibly diminish. Moreover, a pattern stands out for all $T$, on the left side of the dispersion plot, which may suggest a compensatory effect over $k^*$ on the estimation of $\alpha^*$. To evaluate the quality of the estimation procedure, $(\alpha^*|k)$ was computed. The variability in estimates is substantially lower compared to the case of $k^*$, except for $T=1,000$, which appears to lose quality considerably. Thus, suggesting that this sample size is indeed too small to get an accurate $\alpha^*$. Overall, results for $T>1,000$, indicate that, if a reliable estimate of $k$ is provided, then $\alpha$ is accurately estimated.

Table \ref{tab:alpha_stats} shows summary statistics for $(\alpha^*|k^*)$ and $(\alpha^*|k)$ for several sample 
sizes, and aids in clarifying the previous results. The correlation between $(\alpha^*|k)$ for the largest sample size is 0.93, while for $(\alpha^*|k^*)$ is about a third. This can be largely explained by the amount of instances where $\alpha >1$, which are less than 5\% for ($\alpha^*|k$), but over 20\% for ($\alpha^*|k^*$). Also, for ($\alpha^*|k^*$) about 7\% instances return $\alpha > 5$, which further impacts $r$. These extreme lead to more biased estimated of $(\alpha^*|k^*)$, compared to $(\alpha^*|k)$ . The correlation for $T=10,000$ drops to 0.37 for $(\alpha^*|k)$, thus, it is not surprising that for the case $(\alpha^*|k^*)$ this is less than 0.10. For $T=1,000$ both cases have a similar correlation, and estimates are extremely biased, further supporting the fact that this sample size is too small to generate accurate estimates of $\alpha$. 

\begin{table}[]
\caption{Summary statistics for the estimation procedure of  the smoothing factor conditioned to the optimal context $(\alpha^*|k^*)$, and conditioned to the data-generating context $(\alpha^*|k)$ for different sample sizes $T$.}
\centering
\begin{tabular}{@{}lccc@{}}
\toprule
Statistic & $T=100,000$ & $T=10,000$ & $T=1,000$ \\ \midrule
$r(\alpha^*|k^*, \alpha)$ & 0.32 & 0.08 & 0.07 \\
$r(\alpha^*|k, \alpha)$ & 0.93 & 0.37 & 0.07 \\
Bias $\alpha^*|k^*$ & 1.06 & $7.6\times10^{9}$ & $1.44\times10^{11}$ \\
Bias $\alpha^*|k$ & 0.01 & 0.06 & $2.1\times10^{10}$ \\
\% $(\alpha^*|k^*) > 1$ &  22.5&  41.5&  52.3\\
\% $(\alpha^*|k ) > 1$ &  4.3&  8.4&  26.1\\
\% $(\alpha^*|k^*) > 5$ & 7.7  &  23.1 & 33.6 \\ 
\% $(\alpha^*|k) > 5$ & 0 & 0.2 & 6.1  \\ \bottomrule
\end{tabular}
\label{tab:alpha_stats}
\end{table}

\begin{figure}[H]
    \centering
    \begin{subfigure}[b]{0.49\textwidth}
        \centering
        \caption{}
        \includegraphics[width=\linewidth]{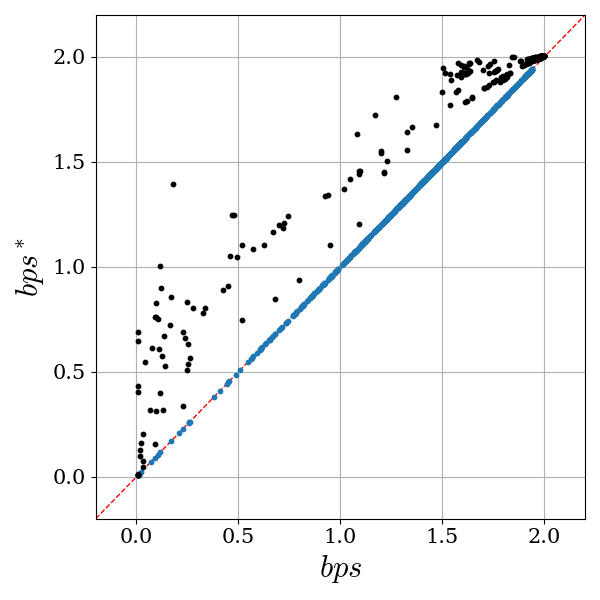}
        \label{A}
    \end{subfigure}
    \begin{subfigure}[b]{0.49\textwidth}
        \centering
        \caption{}
        \includegraphics[width=\linewidth]{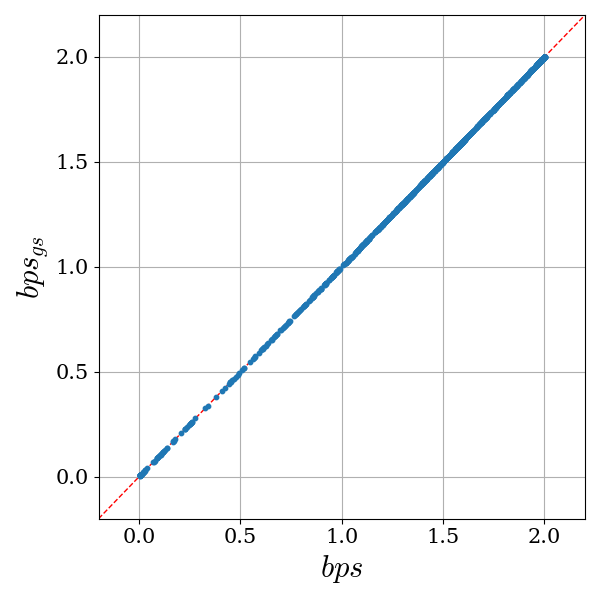}
        \label{B}
    \end{subfigure}
    \caption{
    Dispersion diagrams of $bps$ versus the estimated $bps$. (a) $bps^*$ obtained when compressing with optimal $(k^*,\alpha^*)$ using the proposed approach, distinguishing $k^* = k$ (blue) or $k^* \neq k$ (black). (b) $bps_{gs}$ obtained when compressing with the optimal $(k,\alpha)$ found via grid search. Red line is $r=1$.}
    \label{fig:scatter_bps}
\end{figure}

Figure \ref{fig:scatter_bps} evaluates the impact of the proposed sequential hyperparameter selection procedure on compression performance. The dispersion plots compare the bitrate obtained from the optimal $(k^*,\alpha^*)$, $bps^*$, and bitrate obtained using grid search, $bps_{gs}$ with the $bps$ obtained from the data-generating parameters $(k,\alpha)$.Most points lie close to the diagonal line, indicating that the bitrate obtained with $(k^*,\alpha^*)$ is generally very similar to the bitrate computed with the data-generating process ($k, \alpha$) (Fig. \ref{A}). When the context length is correctly identified, i.e., $k^* = k$, (blue points), the achieved $bps^*$ is indistinguishable from $bps$. In contrast, when the context is misidentified, i.e., $k^* \neq k$, (black points), the resulting compression performance is consistently worse than the optimal one, highlighting the importance of accurately estimating $k$. However, points over the line account for 5\% of the misclassified $k^*$. Thus, suggesting that $\alpha^*$ can, to some extent, compensate for inaccuracies in the estimation of $k$. The bitrate obtained from the grid search ($bps_{gs}$) closely matches the $bps$ obtained with ($k, \alpha$), with the points concentrated along the diagonal. It is evident that there is a perfect correlation between $bps_{gs}$ and $bps$ (Fig. \ref{B}). However, while $bps^*$ is computed with a simple two-step procedure and only needs to compress the data once, $bps_{gs}$ is the result of an exhaustive search procedure over 1,010 points, thus requiring the compressor to be executed  1,010 times.

\section{Conclusion}

This work introduced a two-step sequential approach for the selection of FCM hyperparameters $(k,\alpha)$. The proposed framework decomposes the joint optimization of the context length $k$ and smoothing parameter $\alpha$ into two independent stages. First, $k$ is estimated using  the serial dependence categorical features. Then, conditional on the selected $k^*$, $\alpha^*$ is estimated via maximum likelihood.

The simulation results provide several important insights. First, the experiments show that pami exhibits a clear and consistent pattern that allows to identify $k$, through a maximum value criteria. Thus, pami is clearly superior to the other metrics evaluated, and was capable to successfully identify $k$ in about 70\% of cases for $T=100,000$. Secondly, the estimation of $\alpha$ via maximum likelihood proved effective when $k$ is correctly identified. The accuracy of the estimates increases with the sample size, while small sample sizes lead to larger variability in the estimates. The results also suggest that misidentifying $k$ can propagate to the estimation of $\alpha$, highlighting the importance of a reliable $k$ detection. Lastly, the proposed strategy was evaluated in terms of compression performance. Results show that the bitrate obtained using the estimated hyperparameter $(k^*,\alpha^*)$ is comparable to that obtained through grid search. More importantly, the proposed approach achieves this performance while requiring only a single compression run, whereas grid search requires multiple compressions with a large number of hyperparameter combinations.

Overall these results suggest that the context length $k$ is the dominant hyperparameter in compression efficiency. Thus, next stages of the work should focus on developing other criteria that improve the classification success of $k$. Furthermore, more extensive studies for other alphabet sizes and deeper contexts are also necessary.

\begin{credits}
\subsubsection{\ackname} 
This work was supported by the Foundation for Science and Technology (FCT) through Institute of Electronics and Informatics Engineering of Aveiro (IEETA) contract doi.org/10.54499/UID/00127/2025 and Project “Agenda ILLIANCE” [C644919832-00000035 | Project nº 46], PRR – Plano de Recuperação e Resiliência under the Next Generation EU from the European Union. 

\subsubsection{\discintname}
The authors have no interests to disclose.
\end{credits}
%
%
%
\bibliographystyle{splncs04}
\bibliography{mybibliography}
%




\end{document}